\documentclass[11pt,a4paper]{article}
\usepackage[hyperref]{ranlp2021}
\usepackage{times}
\usepackage{latexsym}

\usepackage{enumitem}

\usepackage{microtype}
\usepackage{placeins}
\usepackage{stfloats}
\usepackage{multirow}

\usepackage{xr-hyper}
\makeatletter
\newcommand*{\addFileDependency}[1]{
  \typeout{(#1)}
  \@addtofilelist{#1}
  \IfFileExists{#1}{}{\typeout{No file #1.}}
}
\makeatother

\newcommand*{\myexternaldocument}[1]{
    \externaldocument{#1}
    \addFileDependency{#1.tex}
    \addFileDependency{#1.aux}
}
\myexternaldocument{supplement}

\aclfinalcopy 

\usepackage{times}
\usepackage{latexsym}
\usepackage{graphicx}
\usepackage{todonotes}
\usepackage{amsmath}
\usepackage{color, colortbl}
\definecolor{Gray}{gray}{0.95}
\usepackage[first=0,last=9]{lcg}

\usepackage{amssymb}

\usepackage{stfloats}
\usepackage{float}
\restylefloat{table}
\usepackage{siunitx,booktabs,subcaption,tabularx}
\usepackage{pbox}

\makeatletter
\DeclareRobustCommand{\textsupsub}[2]{{%
  \m@th\ensuremath{%
    ^{\mbox{\fontsize\sf@size\z@#1}}%
    _{\mbox{\fontsize\sf@size\z@#2}}%
  }%
}}
\makeatother

\title{On the Usability of Transformers-based models for a French Question-Answering task}

\author{Oralie Cattan$^{1,2}$ \And
  Christophe Servan$^{1}$ \And
  Sophie Rosset$^{2}$ \AND \vspace*{-0.5cm}\\ 
  $^{1}$QWANT \\
61 rue de Villiers,\\
92200 Neuilly-sur-Seine, France \\
  \texttt{inital.lastname@qwant.com} \\
  \And \vspace*{-0.5cm}\\
  $^{2}$Université Paris-Saclay, \\CNRS, LISN, \\
  91405, Orsay, France \\
  \texttt{lastname@lisn.fr} \\}

\date{}

\begin{document}

\maketitle

\begin{abstract}


For many tasks, state-of-the-art results have been achieved with Transformer-based architectures, resulting in a paradigmatic shift in practices from the use of task-specific architectures to the fine-tuning of pre-trained language models.
The ongoing trend consists in training models with an ever-increasing amount of data and parameters, which requires considerable resources.
It leads to a strong search to improve resource efficiency based on algorithmic and hardware improvements evaluated only for English. 
This raises questions about their usability when applied to small-scale learning problems, for which a limited amount of training data is available, especially for under-resourced languages tasks.
The lack of appropriately sized corpora is a hindrance to applying data-driven and transfer learning-based approaches with strong instability cases.
In this paper, we establish a state-of-the-art of the efforts dedicated to the usability of Transformer-based models and propose to evaluate these improvements on the question-answering performances of French language which have few resources.
We address the instability relating to data scarcity by investigating various training strategies with data augmentation, hyperparameters optimization and cross-lingual transfer.
We also introduce a new compact model for French FrALBERT which proves to be competitive in low-resource settings. 

\end{abstract}

\section{Introduction}

Recent advances in the field of Natural Language Processing (NLP) have been made with the development of transfer learning and the availability of pre-trained language models based on Transformer architectures \cite{NIPS2017_3f5ee243}, such as BERT \cite{devlin-etal-2019-bert}.
As they provide contextualized semantic representation they contribute both to advance the state-of-the-art on several NLP tasks and also to evolve training practices through the use of fine-tuning.

The trend of recent years consists in training large pre-trained language models on ever larger corpora, with an ever-increasing amount of parameters, which requires considerable computational resources that only a few companies and institutions can afford. 
For example, the base model of BERT with 110 million parameters was pre-trained on 16 gigabytes (GB) of text, while the GPT-3 model \cite{NEURIPS2020_1457c0d6} was pre-trained on 45 terabytes (TB) of text and has 175 billion parameters.

In fact, deploying ever-larger models raises questions and concerns about the increasing magnitude of the temporal, financial, and environmental cost of training and usability \cite{DBLP:conf/acl/StrubellGM19, sustainlp-2020-sustainlp}. 
Typically, due to their resource requirements, these models are trained and deployed for industrial operations on remote servers. 
This leads to a high use of over-the-air communications, which are particularly resource-intensive \cite{8839651}.
In particular, some NLP applications (speech recognition, speech to text, etc.) have some known problems related to network latency, transmission path difficulties, or privacy concerns.
To reduce the impact of these communications, there is a solution that is to allow these models to run directly on peripheral or mobile devices, that is, in environments with limited resources that require lightweight, responsive models and energy efficiency.
Reducing the size of the models is therefore one of the increasingly favoured avenues, especially for the reduction of memory resources and computation time involved in training and use.

To meet these constraints, compact models represent one of the most promising solutions.
As far as we know, they have only been evaluated on the comprehension tasks covered by GLUE \cite{wang-etal-2018-glue} and the question-answering task with the SQuAD corpus \cite{rajpurkar-etal-2016-squad} with abundant data, in English. 
The improvements resulting from the algorithmic optimizations of the models, although significant, raise questions about their effectiveness on lower-scale learning problems on poorly endowed languages.
The works of \citet{zhang2021revisiting} and \citet{mosbach2021on} have furthermore shown degraded performance in these conditions.
These two reflections are at the origin of a double question which our contribution attempts to answer. 
On the one hand, what is the behavior of a Transformer-based model in the context of a question-answering task in French, a task that is poorly endowed in this language?
On the other hand, what are the impacts of algorithmic improvements of these same models in this context?

To answer these questions, we first establish in section \ref{sec:util} a state-of-the-art that is meant to be broad enough to have a shallow overview depicting the ins and outs and issues around the usability of Transformer-based models whose breadcrumb trail is the issue of resources. 
Then, we present in the section \ref{sec:qa} the recent progress of the question-answering task, through the use of these latest models.
In sections \ref{sec:model} and \ref{sec:XP} we introduce our model and present our experiments on the usability of Transformers models in a question-answering task for French, on \textit{FQuAD} \cite{dhoffschmidt-etal-2020-fquad} and \textit{PIAF} \cite{keraron-EtAl:2020:LREC} corpora. 
We propose to address the instability relating to data scarcity by investigating various training strategies with data augmentation, hyperparameters optimization and cross-lingual transfer.
Finally, we present a new compact model for French based on ALBERT \cite{DBLP:conf/iclr/LanCGGSS20}\footnote{Available at HuggingFace's model hub page.}, and compare it to existing monolingual and multilingual models, large and compact, under constrained conditions (notably on learning data).

\section{Usability of Transformers}
\label{sec:util}

In this section we present the ins and out of the Transformer models to understand how the approaches meet the need for better usability.

\subsection{Architecture and pre-trained models}
\label{sec:issues}

The Transformer architecture \cite{NIPS2017_3f5ee243} is based on a stack of encoder-decoder blocks, composed at a high level of forward propagation networks and multi-headed self-attention operations. 
The self-attention layer is the core element of its architecture that enables its efficiency in modeling the semantic context interdependencies between the units or sub-units of the input sequence. 

Transformer-based language models such as BERT \cite{devlin-etal-2019-bert} are pre-trained on large-scale data collections sourced from Wikipedia or Common Crawl (CC) with one or multiple training objectives (masked language modeling, next sentence or sentence order prediction).
This pre-training can be followed by supervised fine-tuning according to the tasks, whether generatives (machine translation, abstractive summarization) or discriminatives (classification, question-answering).
The ensuing fine-tuning phase allows for better initialization of the models parameters while requiring less task-specific data so as to make the training of subsequent tasks faster.

Recently, \citet{zhang2021revisiting} and \citet{mosbach2021on} have nevertheless shown that the commonly adopted practices (the number of iterations, the choice of model layers) when fine-tuning Transformers-based langage models are inappropriate under resource constrained conditions and adversely affect the stability of models performances as overfitting, label noise memorization or catastrophic forgetting.
Added to this, because the pre-training process is particularly constraining, various works have been oriented towards the research and training of efficient models, both in terms of available capacities and resources and in terms of environmental footprint.

\subsection{A search for efficiency}

Reducing the cost of training Transformers-based models has become an active research area.
To this end, methods based on compression techniques or on architecture improvements have been introduced in order to build compact models with comparable performances to large models.

Many works address the issue of model compression with quantization, pruning, knowledge distillation or a combination of these approaches.
The idea of quantization \cite{DBLP:conf/aaai/ShenDYMYGMK20} is to take advantage of the use of lower precision bit-width floats to reduce memory usage and increase computational density. 
Following the same objective, pruning \cite{NEURIPS2019_2c601ad9} consists in removing parts of a model (weight bindings, attentional heads) with minimal precision losses.
Finally, knowledge distillation \cite{DBLP:journals/corr/abs-1910-01108} enables the generation of models that mimic the performance of a large model (or set of models) while having fewer parameters.

Another axis of development concerns the use of neural architecture search \cite{JMLR:v20:18-598} which allows to optimize a model by progressively modifying the design of the network through trial and error, eliminating insignificant operations. 
To avoid the unnecessary large number of parameters, adapters \cite{conf/icml/HoulsbyGJMLGAG19} were introduced to allow fine-tuning of the set of parameters specific to the task of interest rather than the entire model.

Other architectural improvements highlighted with the introduction of the ALBERT model \cite{DBLP:conf/iclr/LanCGGSS20} such as the factorization of the attention matrix or parameter sharing.
Indeed, the most time-consuming and memory-intensive operations concerns the forward propagation and attention computation operations.
The self-attention layer of BERT pretrained models grows quadratically in respect to the input sequence length. 
One common approach to this issue consists of approximating the dot-product attention for example by using hashing techniques \cite{Kitaev2020Reformer} to accelerate the training and inference phases when long sequence lengths are used.
However these solutions have demonstrated they suffer from important computational overheads for tasks with smaller lengths, such as question-answering.


\section{The Question-Answering task}
\label{sec:qa}
Question-Answering (QA) based on machine reading comprehension corresponds to the task of extracting an answer given a question and a context document such as from a news or Wikipedia article.

\subsection{General QA Architecture}
Until recently, most of the proposed approaches have relied on an architectural complexification of LSTM-based neural networks and attention mechanism.
At a high level, their architectures are all composed of three layers with: 

\begin{enumerate}[label=(\alph*)]
    \item \textbf{an encoding layer} that projects the inputs, as each word within the context, question and answer triples in a latent semantic space;
    \item \textbf{an interaction layer} that models the semantic interdependencies between the embedded inputs through the use of attention mechanisms. 
    \item \textbf{an output layer} that extracts the answer to the input question within the related context.
\end{enumerate}

The interaction layer is the core element of the architecture for which several kinds attention mechanisms has been developed to improve the QA matching process such as bi-attention \cite{DBLP:conf/iclr/SeoKFH17}, co-attention \cite{DBLP:conf/iclr/XiongZS17, xiong2018dcn}, multi-level inter-attention \cite{DBLP:conf/iclr/HuangZSC18} or re-attention \cite{DBLP:conf/ijcai/HuPHQW018}, to name just a few.

Recent advances through the availability of Transformer-based pre-trained models and the development of transfer learning methods have enabled to remove the recurrence of previous architectures in order to achieve parallelization efficiencies.
This simplified the QA architecture and its training process, replacing the encoding and the interaction layers with attention-based Transformer layers.

Another advantage is that it provides pre-computed contextual word representations.
QA models based on LSTMs are built on top of static word embeddings models such as GloVe \cite{pennington2014glove}.
Even these models have up to 40 times fewer parameters than a BERT-based model, they rely on LSTM-based encoders to produce contextual embeddings which considerably lengthens the time required for training and makes the dependence on supervised data more important.

The standard approach introduced by \citet{devlin-etal-2019-bert} we rely on this study, consists in introducing and updating parameter vectors corresponding to the start and end positions of the answer span.
Specifically, the start and end position probability distributions are computed by softmax over the dot products between the representation of the tokens and the start and end vectors.
In sum, all of the Transformer parameters as well as the two introduced parameter vectors are optimized together. 

Despite the fact that there is a number of large-scale QA datasets in English, with tens of thousands of annotated training examples, porting a system to a new language with fewer annotated resources (low-resource languages) requires approaches that go far beyond the simple act of retraining the models.

\subsection{Low-Resourced QA}

In recent years, low-resource NLP has drawn an increasing amount of attention with solutions ranging from developing new data collection methodologies either via crowdsourcing or through the use of machine translation (MT), to cross-lingual and transfer learning approaches for which information is shared across languages or tasks.

\subsubsection{MT-based data collection}
\label{sec:mt}
Neural MT as made considerable progress in recent years such as translating large-scale datasets from a high-resourced to under-resourced languages or converserly has become an intuitive way of generating annotated datasets in a cost-effective and rapid manner.

Automatically translating the context, question and answer triples from a high-resource language, such as English (called source domain) to low-resource languages (called target domains) have enabled the evaluation of models for languages with no training data available but also the creation of large-scale MT-based QA corpora for the Italian \cite{10.1007/978-3-030-03840-3_29}, Spanish \cite{carrino-etal-2020-automatic}, Arabic \cite{mozannar-etal-2019-neural} and Korean \cite{NODE09353166} languages. 

Another approach consists of translating the QA triples of the target domain into the source domain, so the model trained on the source language can be directly applied on the translated target language testing data.
As an exemple, \citet{DBLP:journals/corr/abs-1809-03275}'s method consisted of combining the alignment attention scores from a MT model with an English QA model to guide the answer extraction process.

The performance of MT-data approaches depends strongly on the quality of the MT models. 
Thus, due to the lack of reliable models for some language pairs, approaches that foster the transfer of knowledge from other languages or tasks while requiring less data have been developed.

\subsubsection{Pre-training and Transfer approaches}

The exploitation of pre-trained models followed by task-specific fine-tuning haved pushed the state-of-the-art forwards, while requiring much less computational and data resources.
The idea behind pre-training is to reuse the weights parameters trained on a set of source tasks and continue to fine-tune them on under-resourced target tasks to achieve knowledge transfer. 
\citet{NIPS2015_7137debd} were the first to propose to pre-train RNNs using auto-encoders and language models as part of their QA encoding layer. 
\citet{min-etal-2017-question} and \citet{wiese-etal-2017-neural-domain} pre-trained QA models before applying the fine-tuning process between the source and the target domains.
Other efforts focused on pre-training Transformer-based models multilingually such as the multilingual version of BERT (called mBERT) \cite{devlin-etal-2019-bert} or XLM-R \cite{conneau-etal-2020-unsupervised} to learn cross-lingual representations which are transferable across languages.

\subsubsection{Usability concerns}

Studies on the usability of Transformer-based models (Section \ref{sec:util}) from standard resource efficiency concerns towards a broader set of problems related to their generalizability.

Recently, \citet{pires-etal-2019-multilingual} and \citet{conneau-etal-2020-unsupervised} have shown that multilingual models underperformed, when applied on poorly endowed languages.
Additionally, as mentioned in subsection \ref{sec:issues}, recent works \cite{zhang2021revisiting, mosbach2021on} have highlighted the limitation of Transformer-based transfer learning with strong instabilities arising from the small-scale learning.

French is a poorly endowed language since we do not have enough annotated data to train a deep learning model on QA tasks.
Moreover, unlike the only two large monolingual French models: CamemBERT \cite{martin-etal-2020-camembert} and FlauBERT \cite{le-etal-2020-flaubert-unsupervised}, the English BERT model has become a branching point from which a growing number of large and compact English pre-trained models have emerged.
These French monolingual models, although they provide good performances, do not reflect the rapid evolution of the field.

Consequently, in this paper we propose a new compact model FrALBERT for French we present in the following section alongside with the other available pre-trained language models on which we base our experiments.

\section{FrALBERT and Transformer models considered}
\label{sec:model}
As mentioned in the previous section there is no compact model for French. 
We therefore decided to pre-train a new version of ALBERT from scratch we called FrALBERT, thus overcoming some of the discussed limitations.

ALBERT is based on parameter sharing/reduction techniques that allows to reduce the computational complexity and speed up training and inference phases.
Compared to previous compact models such as DistilBERT \cite{DBLP:journals/corr/abs-1910-01108}, Q-BERT \cite{DBLP:conf/aaai/ShenDYMYGMK20} or TernaryBERT \cite{zhang-etal-2020-ternarybert}, ALBERT is to the date the smallest pre-trained models with 12 million parameters and \textless50 megabyte (MB) model size.


FrALBERT is pre-trained on the French version of the Wikipedia encyclopedia of 04/05/2021, i.e. 4 GB of text and 17 million (M) sentences. 
Beyond concerns about the rights to use data from Common Crawl projects such as OSCAR \cite{ortiz-suarez-etal-2020-monolingual} or CCNet \cite{wenzek2020ccnet} corpora, and because we focus on factual QA, we decide to use only Wikipedia as our primary source of knowledge.
We used the same learning configuration as the original model with a batch size of $128$ and a initial learning rate set to $3.125\times10\textsuperscript{-4}$.

Our experiments are also based on the large monolingual French model CamemBERT \cite{martin-etal-2020-camembert} as well as on the two large multilingual models: XLM-R \cite{conneau-etal-2020-unsupervised} and mBERT \cite{devlin-etal-2019-bert}, both pre-trained from massive corpora dataset in more than 100 languages such as the Common Crawl (CC-100) or Wikipedia (Wiki-100).
We also exploit two compact multilingual models with a distilled version of mBERT: distil-mBERT \cite{DBLP:journals/corr/abs-1910-01108} and small-mBERT \cite{abdaoui-etal-2020-load}, a mBERT model whose the original vocabulary has been reduced to two languages (English and French). 
Table \ref{table:param} gives a comparison of the models.



\begin{table*}[ht]
\small
\centering
\scalebox{1}{
\begin{tabular}{llrrr}
\hline
\textbf{model} & \multicolumn{1}{c}{\textbf{pre-training data}} & \multicolumn{1}{c}{\textbf{vocab. size}} & \multicolumn{1}{c}{\textbf{\# param.}} & \multicolumn{1}{c}{\textbf{model size}} \\ \hline
\textbf{CamemBERT}\textsubscript{base}  & French OSCAR (138 GB of text)      &  32005           & 110 M                          & 445 MB                           \\
\rowcolor{Gray} \textbf{CamemBERT}\textsubscript{large}  & French CCNet (135 GB of text)  &  32005           & 335 M                          & 1.35 GB                          \\
\textbf{CamemBERT}\textsubscript{base}  & French Wikipedia (4 GB of text) &    32005         & 110 M                          & 445 MB                           \\
\rowcolor{Gray} \textbf{FrALBERT}\textsubscript{base}  & French Wikipedia (4 GB of text) &    32005         & 12 M                           & 50 MB                            \\
\textbf{XLM-R}\textsubscript{base}   & CC-100 (2.5 TB of text)           &     250002        & 278 M                          & 1.12 GB                          \\
\rowcolor{Gray}\textbf{XLM-R}\textsubscript{large}   & CC-100 (2.5 TB of text)           &      250002       & 559 M                          & 1.24 GB                          \\
\textbf{mBERT}\textsubscript{base}   & Wiki-100         &     119547        & 177 M                          & 714 MB                        \\
\rowcolor{Gray} \textbf{small-mBERT}\textsubscript{base}  & Wiki-100         &   33407          & 111 M                          & 447 MB                           \\
\textbf{distil-mBERT}\textsubscript{base}  & Wiki-100         &     119547        & 134 M                          & 542 MB                           \\ \hline
\end{tabular}
}
\caption{Characteristics of the pre-trained models used in the proposed small-scale QA framework.}
\label{table:param}
\end{table*}

\begin{table*}[hb]
\small
\centering
\scalebox{1}{
\begin{tabular}{lcccc}
\hline
\textbf{dataset}                                 & \textbf{\textit{SQuAD-en}\textsubscript{train}}   & \textbf{\textit{FQuAD}\textsubscript{train}} & \textbf{\textit{FQuAD}\textsubscript{dev}}      & \textbf{\textit{PIAF}\textsubscript{dev}}       \\ \hline
\textbf{\# titles}                               & 442                 & 117                 & 18                  & 191                 \\ \hline
\textbf{\# paragraphs}                           & 18,891              & 4,920                & 768                 & 761                 \\
\textbf{\# sentences / \# tokens / \#characters} & 5.0 / 119.7 / 635.6 & 4.8 / 125.9 / 653.8 & 5.4 / 147.3 / 765.7 & 4.2 / 110.5 / 579.1 \\ \hline
\textbf{\# questions}                            & 87,342              & 20,088              & 3,184               & 3,812               \\
\textbf{\# tokens / \# characters}               & 10.1 / 50.2         & 9.2 / 47.9          & 8.5 / 45.6          & 9.1 / 48.2          \\ \hline
\textbf{\# answers}                              & 87,599              & 20,731              & 3,188               & 3,835               \\
\textbf{\# tokens / \# characters}               & 2.9 / 16.8          & 3.6 - 19.5          & 4.3 - 23.0          & 4.5 - 24.1          \\ \hline
\textbf{human performance (F1 / EM)}             & 90.5 / 80.3         & \multicolumn{2}{c}{92.6 / 79.5}          & -                   \\ \hline
\end{tabular}
}
\caption{Descriptives of \textit{SQuAD-en}, \textit{FQuAD} and \textit{PIAF} datasets.}
\label{table:stats}
\end{table*}

\section{Experiments}
\label{sec:XP}

We propose an assessment of the comparative advantage gains in performance when using different training strategies (data augmentation, hyperparameter search and cross-lingual transfer) over monolingual and multilingual pre-trained models, larges and compacts for a QA task in French under resource constraints. 

\subsection{QA Datasets}

We conduct experiments on four QA datasets whose descriptives are presented in Table \ref{table:stats}, with:

\begin{itemize}
    \item \textit{SQuAD} (v1.1) \cite{rajpurkar-etal-2016-squad} (we called \textit{SQuAD-en}) the reference corpus to evaluate QA models' performances in English, consisting of 100K+ QA pairs sourced from 442 English Wikipedia articles;
    \item \textit{FQuAD} (v1.0) \cite{dhoffschmidt-etal-2020-fquad}, a recently released French QA dataset consisting of 25K+ crowdsourced QA pairs based on 135 articles on French Wikipedia;
    \item \textit{PIAF} (v1.0) \cite{keraron-EtAl:2020:LREC}, a small-scale dataset in French with only 3K+ pairs of QA pairs in 191 Wikipedia articles; and 
    \item \textit{SQuAD-fr}\textsubscript{train}, our French translated version of \textit{SQuAD-en}. 
    We used the Transformer architecture as described in \citet{NIPS2017_3f5ee243} from the Open NMT framework \cite{klein-etal-2017-opennmt} (Open-Source Neural Machine Translation) implementation of the network to train our neural MT system. When translating QA corpora, the problem we face is that the translated answer may not be present in the translated context. Thus, simple techniques such as segment matching are inadequate to retrieve the answer. We have developed an answer extraction process that is based on ChrF \cite{popovic-2015-chrf} a character n-gram precision and recall enhanced with word n-grams.  Since answers are largely made up of entities, ChrF score integration is only performed when the answer span is not present in the related context. 
    In order to evaluate the quality of the translation, we manually corrected the translation errors in the output of a subset of the corpus composed of 890 QA pairs and 107 contexts. 
    We obtain a BLEU score \cite{papineni2002bleu} of 68.89 and 72.38 for questions and contexts respectively.
    \textit{SQuAD-fr}\textsubscript{train} serves as a means of data augmentation on \textit{FQuAD} and \textit{PIAF} benchmarks, with 90K+ translated QA training pairs\footnote{We share our \textit{SQuAD-fr} corpus on request and on dataset sharing platforms to support further research in this area.}.
\end{itemize}

We also explore mixed datasets training strategy with \textit{SQuAD-en}\textsubscript{train} $+$ \textit{FQuAD}\textsubscript{train} for training models on a concatenation of the training data covering French-English language pairs to test the cross-lingual transfer ability of multilingual models.

\subsection{Evaluation and validation}

The performance of QA models are evaluated using the Exact Match (EM) and F1 scores. 
The EM score is the percentage of system outputs that match exactly with the ground truth answers. The F1 score is a combined measure of precision and recall that is less strict than EM.
The evaluation process\footnote{The experiments reported in \cite{dhoffschmidt-etal-2020-fquad} concern version 1.1 of the \textit{FQuAD} corpus. Ours are based on the only version available to date (v1.0). Moreover, the test sets are not made public, so we use the development set instead.} involves post-processing identical to that presented by \citet{dhoffschmidt-etal-2020-fquad} and inspired by that proposed for English by \citet{rajpurkar-etal-2016-squad}, which consists of the removal of punctuation marks and determiners\footnote{Determiners are \textit{le, la, les, l', du, des, au, aux, un, une}.} as well as a down-casing of the answers (ground truths and predictions). 

To address our considerations related to resource constraints we perform a hyperparameter optimization, that has proven to lead to better solutions in less time.
It is based on a population-based learning \cite{jaderberg2017population} in which a population of models and their hyperparameters are jointly optimized. To this end, we build a validation set by randomly extracting 10\% of the training data.

\begin{table*}[ht!]
\small 
\resizebox{\textwidth}{!}{%
\begin{tabular}{l|ccc|ccc}
\hline
\textbf{testing data}                                                            & \multicolumn{3}{c|}{\textbf{\textit{FQuAD}\textsubscript{dev}}}                                                                                                                       & \multicolumn{3}{c}{\textbf{\textit{PIAF}\textsubscript{dev}}}                                                                                                                               \\ \hline
\multicolumn{1}{l|}{\textbf{model \textbackslash ~training strategy}} & \textbf{\begin{tabular}[c]{@{}c@{}}\textit{FQuAD}\textsubscript{train}\end{tabular}} & \textbf{\begin{tabular}[c]{@{}c@{}}\textit{FQuAD}\textsubscript{train} \\ w/ optim.\end{tabular}}           & \textbf{\begin{tabular}[c|]{@{}c@{}}\textit{FQuAD}\textsubscript{train} \\ $+$ \textit{SQuAD-fr}\textsubscript{train}\end{tabular}}    & \textbf{\begin{tabular}[c]{@{}c@{}}\textit{FQuAD}\textsubscript{train}\end{tabular}} & \textbf{\begin{tabular}[c]{@{}c@{}}\textit{FQuAD}\textsubscript{train} \\ w/ optim.\end{tabular}} & \textbf{\begin{tabular}[c]{@{}c@{}}\textit{FQuAD}\textsubscript{train} \\ $+$ \textit{SQuAD-fr}\textsubscript{train}\end{tabular}}   \\ \hline 
\multicolumn{1}{l|}{\textbf{CamemBERT}\textsubscript{base}}                         & \multicolumn{1}{c|}{77.6 / 52.5}                              & \multicolumn{1}{c|}{85.5 / 70.3} & \multicolumn{1}{c|}{\textbf{86.7 / 71.7}} & \multicolumn{1}{c|}{62.0 / 37.5}                              & \multicolumn{1}{c|}{63.8 / 38.9}                                 & \textbf{64.3 / 39.2}  \\ 
\rowcolor{Gray}\multicolumn{1}{l|}{\textbf{CamemBERT}\textsubscript{large}}                        & \multicolumn{1}{c|}{81.2 / 55.9}                              & \multicolumn{1}{c|}{\textbf{90.2 / 75.5}} & \multicolumn{1}{c|}{89.9 / 75.2}                        & \multicolumn{1}{c|}{68.1 / 42.2}                              & \multicolumn{1}{c|}{\textbf{71.0 / 44.8}}                                 & 68.9 / 42.5  \\ 

\multicolumn{1}{l|}{\textbf{CamemBERT}\textsubscript{base (wiki 4 GB)}}                    & \multicolumn{1}{c|}{74.2 / 49.5 }                              & \multicolumn{1}{c|}{80.7 / 61.8} & \multicolumn{1}{c|}{\textbf{85.1 / 69.5}} & \multicolumn{1}{c|}{61.7 / 37.3 }                              & \multicolumn{1}{c|}{62.9 / 37.9}                                 & \textbf{65.9  / 41.0} \\ 
\rowcolor{Gray} \multicolumn{1}{l|}{\textbf{FrALBERT}\textsubscript{base (wiki 4 GB)}}                    & \multicolumn{1}{c|}{72.6 / 55.1}                              & \multicolumn{1}{c|}{75.6 / 64.8} & \multicolumn{1}{c|}{\textbf{84.3 / 70.5} } & \multicolumn{1}{c|}{61.0 / 38.9}                              & \multicolumn{1}{c|}{62.1 / 39.5}                                 & \textbf{66.9 / 43.7} \\ 
\multicolumn{1}{l|}{\textbf{XLM-R}\textsubscript{base}}                    & \multicolumn{1}{c|}{82.1 / 66.8}                              & \multicolumn{1}{c|}{83.1 / 67.9} & \multicolumn{1}{c|}{\textbf{84.2 / 68.8}} &
\multicolumn{1}{c|}{65.0 / 39.6}                              & \multicolumn{1}{c|}{66.9 / 41.2}                                 & \textbf{68.6 / 42.7} \\

\rowcolor{Gray}\multicolumn{1}{l|}{\textbf{XLM-R}\textsubscript{large}}                    & \multicolumn{1}{c|}{86.8 / 71.5}                              & \multicolumn{1}{c|}{\textbf{89.5 / 75.8}} & \multicolumn{1}{c|}{87.3 / 72.5} & 
\multicolumn{1}{c|}{70.4 / 43.8}                              & \multicolumn{1}{c|}{\textbf{73.2 / 45.8}}                                 &  72.6 / 45.2                       \\

\multicolumn{1}{l|}{\textbf{mBERT}\textsubscript{base}}                    & \multicolumn{1}{c|}{78.6 / 61.8}                              & \multicolumn{1}{c|}{82.5 / 65.7} & \multicolumn{1}{c|}{\textbf{84.1 / 68.6}} & \multicolumn{1}{c|}{62.5 / 37.8}                              & \multicolumn{1}{c|}{64.1 / 38.0}                              & \textbf{64.8 / 40.0} \\ 
\rowcolor{Gray}\multicolumn{1}{l|}{\textbf{small-mBERT}\textsubscript{base}}                    & \multicolumn{1}{c|}{75.1 / 55.7}                              & \multicolumn{1}{c|}{78.0 / 62.2} & \multicolumn{1}{c|}{\textbf{81.6 / 64.6}} & \multicolumn{1}{c|}{60.8 / 35.6}                              & \multicolumn{1}{c|}{62.2 / 37.7}                                 & \textbf{63.7 / 39.8}  \\ 
\multicolumn{1}{l|}{\textbf{distil-mBERT}\textsubscript{base}}                    & \multicolumn{1}{c|}{72.8 / 56.0}                              & \multicolumn{1}{c|}{73.0 / 55.1} & \multicolumn{1}{c|}{\textbf{78.1 / 61.5}} & \multicolumn{1}{c|}{52.3 / 30.1}                              & \multicolumn{1}{c|}{53.6 / 31.4}                                 & \textbf{58.3 / 34.9}   \\ \hline

\end{tabular}
}
\caption{Results obtained with French and multilingual Transformer models on the baseline training (\textit{FQuAD}\textsubscript{train}), using hyperparameter optimization (\textit{FQuAD}\textsubscript{train} w/ optim) and with data augmentation (\textit{FQuAD}\textsubscript{train} $+$ \textit{SQuAD-fr}\textsubscript{train}) using F1-measure (F1) and Exact Match (EM), on two French QA tasks (\textit{FQuAD}\textsubscript{dev} and \textit{PIAF}\textsubscript{dev}).
}  
\label{table:mono}
\end{table*}

\subsection{Results}


Table \ref{table:mono} presents the results on the French QA task evaluated on \textit{FQuAD}\textsubscript{dev} and \textit{PIAF}\textsubscript{dev}.
This table shows the scores obtained with Transformer-based models on the baseline training (\textit{FQuAD}\textsubscript{train}), using hyperparameter optimization approach (\textit{FQuAD}\textsubscript{train} w/ optim) and with data augmentation  approach (\textit{FQuAD}\textsubscript{train} $+$ \textit{SQuAD-fr}\textsubscript{train}).
Cross-lingual performances are presented in table \ref{table:cross}, on the same French QA tasks \textit{FQuAD} and \textit{PIAF} as baseline the English corpus \textit{SQuAD-en}\textsubscript{train}, on which we applied hyperparameter optimization (\textit{SQuAD-en}\textsubscript{train} w/ optim) and performed data augmentation by adding the French corpus \textit{FQuAD} to the English QA training corpus (\textit{SQuAD-en}\textsubscript{train} $+$ \textit{FQuAD}\textsubscript{train}).



\subsubsection{Baseline results}

The results obtained from monolingual models with CamemBERT\textsubscript{large} which has more layers, hidden units and attention heads and Camembert\textsubscript{base}, both pre-trained with a larger and more diverse amount of data achieve results are better than Camembert\textsubscript{base} pre-trained on only 4 GB Wikipedia. The highest F1 score is 81.2 on \textit{FQuAD}\textsubscript{dev} and 68.1 on \textit{PIAF}\textsubscript{dev}. 

The F1 performances of the FrALBERT\textsubscript{base} model are close to those of the CamemBERT\textsubscript{base} model, both pre-trained on the French content of Wikipedia (4GB). 
Their results turn out to be competitive and of the same order of magnitude as those reported by \citet{DBLP:conf/iclr/LanCGGSS20} on \textit{SQuAD-en} with 1 point difference on F1 scores when evaluating a BERT\textsubscript{base} model (90.4 F1) and a compact ALBERT\textsubscript{base} model (89.3 F1) pre-trained on the same texts (BookCorpus and Wikipedia). 
Interestingly, these EM scores are higher than those of the CamemBERT\textsubscript{base} achieving the EM score of 55.1, an increase of 5 points on the FQuAD\textsubscript{dev}.

\subsubsection{hyperparameter results}

Automatically tuning the hyperparameter tends to make QA models more accurate with gains in terms of EM scores that are very expressive.
Highest F1 / EM scores are 90.2 / 75.5 on \textit{FQuAD}\textsubscript{dev} and 71.0 / 44.8 on \textit{PIAF}\textsubscript{dev}. 
Improvement are variable accoring the model considered, especially the French BERT one wich have the highest improvement using this approach (from 6 to 9 F1 points and from 11 to 20 EM points) which is quite imrpessive. 
FrALBERT stay behind of 5 F1 points of the CamemBERT\textsubscript{base} trained with the same data (wiki 4 GB) but regarding the EM scores, FrALBERT is better of 3 points.
Surprisingly, multilingual models are close to the French BERT models. 
The small-mBERT is better than FrALBERT around 2.5 F1 points, while the French one have a better EM score (+2.6 points), while distil-mBERT is lower in both F1 and EM scores.


\subsubsection{Data augmentation results}

Training strategies based on data augmentation got nearly the best results in both F1 and EM scores except for the CamemBERT\textsubscript{large}. 
Apart from CamemBERT\textsubscript{base (wiki 4 GB)} and FrALBERT\textsubscript{base} models, results are comparable with an average difference of 1 point regardless of the metric.
More generally, the performance gains are up to 11 and 20 of F1 and EM points, respectively, on \textit{FQuAD}\textsubscript{dev} and up to 4 points on both metrics on \textit{PIAF}\textsubscript{dev}. 

\begin{table*}[ht!]
\small 
\resizebox{\textwidth}{!}{%
\begin{tabular}{l|ccc|ccc}
\hline
\textbf{testing data}                                                            & \multicolumn{3}{c|}{\textbf{\textit{FQuAD}\textsubscript{dev}}}                                                                                                                       & \multicolumn{3}{c}{\textbf{\textit{PIAF}\textsubscript{dev}}}                                                                                                                               \\ \hline
\multicolumn{1}{l|}{\textbf{model \textbackslash ~training strategy}} & \textbf{\begin{tabular}[c]{@{}c@{}}\textit{SQuAD-en}\textsubscript{train}\end{tabular}} & \textbf{\begin{tabular}[c]{@{}c@{}}\textit{SQuAD-en}\textsubscript{train} \\ w/ optim.\end{tabular}}           & \textbf{\begin{tabular}[c|]{@{}c@{}}\textit{SQuAD-en}\textsubscript{train} \\ $+$ \textit{FQuAD}\textsubscript{train}\end{tabular}}    & \textbf{\begin{tabular}[c]{@{}c@{}}\textit{SQuAD-en}\textsubscript{train}\end{tabular}} & \textbf{\begin{tabular}[c]{@{}c@{}}\textit{SQuAD-en}\textsubscript{train} \\ w/ optim.\end{tabular}} & \textbf{\begin{tabular}[c]{@{}c@{}}\textit{SQuAD-en}\textsubscript{train} \\ $+$ \textit{FQuAD}\textsubscript{train}\end{tabular}}   \\ \hline 
\multicolumn{1}{l|}{\textbf{XLM-R}\textsubscript{base}}                    & \multicolumn{1}{c|}{81.3 / 65.0}                              & \multicolumn{1}{c|}{82.5 / 66.5} & \multicolumn{1}{c|}{\textbf{83.6 / 67.5}} & \multicolumn{1}{c|}{61.4 / 37.2}                              & \multicolumn{1}{c|}{62.7 / 38.5}                                 & \textbf{64.9 / 39.9}  \\ 
\rowcolor{Gray}\multicolumn{1}{l|}{\textbf{XLM-R}\textsubscript{large}}                    & \multicolumn{1}{c|}{82.8 / 64.8}                              & \multicolumn{1}{c|}{84.4 / 67.8} & \multicolumn{1}{c|}{\textbf{87.1 / 72.0}} & \multicolumn{1}{c|}{65.1 / 39.1}                              & \multicolumn{1}{c|}{66.3 / 40.5}                                 & \textbf{69.0 / 43.2}  \\ 
\multicolumn{1}{l|}{\textbf{mBERT}\textsubscript{base}}                    & \multicolumn{1}{c|}{76.0 / 59.3}                              & \multicolumn{1}{c|}{79.5 / 62.3} & \multicolumn{1}{c|}{\textbf{83.5 / 67.6}} & \multicolumn{1}{c|}{61.6 / 37.2}                              & \multicolumn{1}{c|}{62.1 / 36.9}                               & \textbf{64.5 / 39.6} \\ 
\rowcolor{Gray}\multicolumn{1}{l|}{\textbf{small-mBERT}\textsubscript{base}}                    & \multicolumn{1}{c|}{73.1 / 49.0}                              & \multicolumn{1}{c|}{76.0 / 59.1} & \multicolumn{1}{c|}{\textbf{81.4 / 62.1}} & \multicolumn{1}{c|}{59.6 / 36.5}                              & \multicolumn{1}{c|}{61.0 / 37.8}                                 & \textbf{63.0  / 38.9}  \\ 
\multicolumn{1}{l|}{\textbf{distil-mBERT}\textsubscript{base}}                    & \multicolumn{1}{c|}{65.4 / 47.4}                              & \multicolumn{1}{c|}{68.6 / 48.5} & \multicolumn{1}{c|}{\textbf{75.9 / 56.3}} & \multicolumn{1}{c|}{48.8 / 28.1}                              & \multicolumn{1}{c|}{52.0 / 29.2}                                 & \textbf{56.5 / 33.1}   \\ \hline

\end{tabular}
}
\caption{Cross-language transfer results obtained with multilingual Transformer models only on the baseline (\textit{SQuAD-en}\textsubscript{train}), using hyperparameter optimization (\textit{SQuAD-en}\textsubscript{train} w/ optim) and with data augmentation (\textit{SQuAD-en}\textsubscript{train} $+$ \textit{FQuAD}\textsubscript{train}) using F1-measure (F1) and Exact Match (EM), on two French QA tasks (\textit{FQuAD}\textsubscript{dev} and \textit{PIAF}\textsubscript{dev}).
}  
\label{table:cross}
\end{table*}

\subsubsection{Cross-lingual transfer results}
The cross-lingual transfer-based approaches using multilingual models outperform the monolingual approaches on \textit{FQuAD} and \textit{PIAF} corpora (table \ref{table:cross}).
Once again the large model XLM-R achieves better results than its base version. 
XLM-R pre-trained with a dual LM objective lens scores better than the mBERT model every time. 
The highest F1 score is 86.8 on \textit{FquAD}\textsubscript{dev} and 70.4 on \textit{PIAF}\textsubscript{dev}. 
There is a significant performance drop between the large multilingual models and their respective compact models.
The compact multilingual models based on mBERT substantially underperform, obtaining lower F1 and EM scores than the large models regardless of the training strategy.
In the zero-shot configurations where no French data is used for training (\textit{SQuAD-en}\textsubscript{train} and \textit{SQuAD-en}\textsubscript{train} w/ optim), the models confirm the outstanding crosslingual ability with performances exceeding the performances of the monolingual models with F1 scores with an F1 / EM scores slightly below those obtained on \textit{FQuAD}\textsubscript{train}.

\subsubsection{General observations}

In all configurations, the performance in terms of EM and F1 on PIAF remains significantly lower than that obtained on \textit{FQuAD} since the \textit{PIAF} corpus does not include multiple responses as pointed out by \citet{dhoffschmidt-etal-2020-fquad}.
Unsurprisingly, \textit{PIAF}\textsubscript{dev} offer a more challenging evaluation set, where the answer extraction performance are lower.
Indeed, the corpus is more diversified with questions on 191 different Wikipedia articles, whereas on \textit{FQuAD}\textsubscript{dev} it only covers 18.

According results, we can confirm that data augmentation is the better way to improve results, even if data comes from another language.
We observe from multilingual results that combining training data gives better results with similar performance whether data are translated or not.

\subsection{Analysis}
In this section, we conduct an analysis of our results to understand what remains as challenges for state-of-the-art models, with a focus on the usability concerns of Transformers under resource constraints.

The success of supervised methods depends heavily on the availability of large-scale training data.
Pre-training large models on massive corpora using unsupervised language modeling and fine-tuning the model with pre-trained weights requires less task-specific data.
Our experimental results are in line with this, since we obtain satisfactory results with transfer learning when large and high quality annotated data are not available.
The highest F1 score is 81.2 on \textit{FQuAD}\textsubscript{dev} and 68.1 on \textit{PIAF}\textsubscript{dev}. 
Nevertheless, the performance of the models can benefit from several training strategies.

\paragraph{Effect of MT-data augmentation}
The lack of human-annotated datasets for languages other than English can be overcome by enriching our training data with the translated version of \textit{SQuAD-en}\textsubscript{train}. 
Regardless of the pre-trained model used, their performance are competitive on \textit{FQuAD}\textsubscript{dev} and \textit{PIAF}\textsubscript{dev}, close to human performance.

\paragraph{Effect of hyperparameter tuning} 
A generally unstated assumption is that pre-trained linguistic models are under-optimized and that practices commonly adopted for the fine-tuning stage can be detrimental to performance \cite{zhang2021revisiting, mosbach2021on}.
This is quite apparent in all settings, with better gains through hyperparameter optimization stages.
Fine-tuning CamemBERT\textsubscript{large} on the French dataset yields 90.2 / 75.5 F1 / EM on the \textit{FQuAD} dev set. 
By means of comparison, CamemBERT\textsubscript{large} scores were 81.2 / 55.9 F1 / EM on the same set with no hyperparameter tuning. 

\paragraph{Crosslingual QA}
Pre-training language models on the concatenation of multiple languages has proven to be a competitive approach for cross-lingual language modeling.
If monolingual models often perform better than multilingual models, we observe that, for comparable model sizes this is not the case in our task where the performance gap is smaller. 
This gap is further reduced when the strategies are combined.
Scores of fine tuned XLM-R\textsubscript{base} is 82.1 / 66.8 on \textit{FQuAD}\textsubscript{dev} and 65.0 / 39.6 on \textit{PIAF}\textsubscript{dev}. 

The zero-shot experiments show that multilingual models can reach strong performances on the task in French when the model has not encountered data of the French language.
For example, the XLM-R\textsubscript{base} model fine-tuned solely on SQuAD-en\textsubscript{train} reaches a performance on FQuaD just a few points below the performance obtained when finetuning is performed on FQuAD\textsubscript{train}.

Finally, our results suggest that data-driven augmentation, either by translating datasets from high resource languages or by concatenating the available corpora are a particularly appropriate strategy to exploit the potential of cross-lingual transferability of models and data for improving model performances.

\paragraph{Improvements over a small scale dataset}

With resource-limited training data we obtain an average F1 score of 74.2 on \textit{FQuAD}\textsubscript{dev} and 61.7 on \textit{PIAF}\textsubscript{dev} when we fine-tuned FrALBERT — higher performance than any of the compact multilingual models, but slightly below the performance of the large monolingual models.
We believe that the lower performance of small multilingual models is not due to their lower number of parameters, but to the usability of these models which is dependent on the reduction technique used. This can be seen very clearly since the performance of the FrALBERT model is close to that of the large models.

Here again, these performances can be boosted via the use of translated data or hyperparameter search which allows us to bring the maximum performances obtained with FrALBERT\textsubscript{base} and CamemBERT\textsubscript{base} pre-trained on 4 GB Wikipedia closer in a consistent way.
Their scores remain slightly below those obtained with models pre-trained on more data suggesting limitations related to the corpus domain of the language model. 

\paragraph{Computational costs}

Compact models provide alternatives to high-energy consumption models by showing comparable performance while reducing their size and computational complexity. Decreasing the environmental impact of NLP model training, as a research topic, is very recent \cite{sustainlp-2020-sustainlp}.
We decided to monitor our experiments conducted on one NVIDIA V100 GPU with 16GB of memory using \texttt{experiment-impact-tracker} \cite{JMLR:v21:20-312}.
Table \ref{table:cost} shows the energy consumption in kilowatt-hour (kWh), the emission intensity in grams of carbon dioxide (g CO$_2$), and the duration in seconds ($s$) for a fine tuning session of 10 epochs with a batch size of 4 on \textit{FQuAD}\textsubscript{train}.

The footprint of the large versions of XLM-R\textsubscript{large} and CamemBERT\textsubscript{large} models is 3 times more than their base versions. Their training time is also significantly longer, over 5 hours.
Finally, in terms of watt usage, carbon emissions and training time, FrALBERT is two times less the distilled version of BERT.

\begin{table*}[ht]
\small
\centering
\scalebox{1}{
\begin{tabular}{lrrrcc}
\hline
\textbf{model} & \multicolumn{1}{c}{\textbf{\# param.}} 
& \multicolumn{1}{c}{\textbf{model size}} 
& \textbf{Time (s)} 
& \textbf{Energy (kWh)} 
& \textbf{CO$_2$ (g)} 
\\ \hline
\textbf{CamemBERT}\textsubscript{base}  & 110 M                          & 445 MB  & 7,207  & 1.08  & 317.87                  \\
\rowcolor{Gray} \textbf{CamemBERT}\textsubscript{large}  &  335 M                          & 1.35 GB      & 19,445  & 3.10  & 914.27                   \\
\textbf{FrALBERT}\textsubscript{base}  & \textbf{12 M}                           & \textbf{50 MB}        & \textbf{3,816}  & \textbf{0.57}  & \textbf{167.80}              \\ \hline

\rowcolor{Gray} \textbf{XLM-R}\textsubscript{base}   & 278 M                          & 1.12 GB       & 7,676  & 1.14  & 337.70             \\
\textbf{XLM-R}\textsubscript{large}   & 559 M                          & 1.24 GB         & 21,137 & 3.30   & 973.29            \\
\rowcolor{Gray} \textbf{mBERT}\textsubscript{base}   &  177 M                          & 714 MB        & 7,333 & 1.07   & 317.02          \\
 \textbf{small-mBERT}\textsubscript{base}  & 111 M                          & 447 MB          & 7,190  & 1.09   & 321.42         \\
\rowcolor{Gray} \textbf{distil-mBERT}\textsubscript{base}  & 134 M                          & 542 MB         & 6,466  & 1.06   & 314.17               \\ \hline
\end{tabular}
}
\caption{Comparison of models by computational costs on \textit{FQuAD}\textsubscript{train}}
\label{table:cost}
\end{table*}

\section{Conclusion and outlook}

Recently, important progress has been made in neural language modeling using Transformer networks.
Its popularity now well established lies in its effectiveness in modeling long-term dependencies.
In this study, we have shown that a number of significant shortcomings of usability have recently been pointed out and that some solutions have been drawn up with compact models. 
We have also overviewed how the use of Transformer-based pre-trained language models have sparked a paradigmatic shift in question-answering training practices from task-specific architectures to the use of transfer learning through fine-tuning.

Comparing performances on a French question-answering task using large and compact models provides insight into the usability of these models for under-resourced languages. 
As others, we argued that large and compact models cannot be used with limited data.
Our experimental results suggest that training strategy such as hyperparameter tuning or data augmentation can help to alleviate the data-gathering burden, with performances close to those of a high-resourced language such as English.

Finally, we present a new compact model for French FrALBERT (12M parameters), which proves to be as competitive as the large monolingual model CamemBERT (110M parameters) pre-trained on the same amount of text. 
In term of computational cost, we shown this compact model is twice less greedy than the BERT\textsubscript{base} models.
We also release a high-quality translated version of the SQuAD corpus in French consisting of around 90K+ QA pairs.

In a future work, we aim to continue this study from a meta-learning perspective with a model-agnostic approach generalizable to low-resource languages.
We also plan to extend our model to other languages and to evaluate it on other NLP tasks such as named entity recognition or natural language understanding.

\section*{Acknowledgments}
This work has been partially funded by the French National Research Agency (ANR) through the project TextToKids (AAPG2019).

\bibliographystyle{acl_natbib}
\bibliography{anthology, ranlp2021}


\end{document}